\documentclass[twocolumn]{article}
 
\usepackage{arxiv}
\usepackage[utf8]{inputenc}
\usepackage{amsmath}
\usepackage{comment}
\usepackage{bm}  
\usepackage{graphicx}
\usepackage{enumitem}
\usepackage[linesnumbered,algoruled,boxed,lined]{algorithm2e}
\usepackage[export]{adjustbox}
\usepackage[table,xcdraw]{xcolor}
\usepackage[pagebackref=true,breaklinks=true,colorlinks,bookmarks=false]{hyperref}    

\usepackage{subcaption}
\usepackage{booktabs}

\begin{document}

\title{Mitigating the Bias of Centered Objects in Common Datasets}

\author{Gergely Szabó, András Horváth\\
Peter Pazmany Catholic University Faculty of Information Technology and Bionics\\
Budapest, Práter u. 50/A, 1083\\
{\tt\small szabo.gergely, horvath.andras @itk.ppke.hu}
}

\twocolumn[
\begin{@twocolumnfalse}
\maketitle
\end{@twocolumnfalse}
]

\section*{Abstract}
Convolutional networks are considered shift invariant, but it was demonstrated that their response may vary according to the exact location of the objects.
In this paper we will demonstrate that most commonly investigated datasets have a bias, where objects are over-represented at the center of the image during training. This bias and the boundary condition of these networks can have a significant effect on the performance of these architectures and their accuracy drops significantly as an object approaches the boundary.
We will also demonstrate how this effect can be mitigated with data augmentation techniques and architectural changes.

\section{Introduction}

Evaluation and validation of machine learning tools in the image processing field is frequently performed on large, publicly available data sets such as the MNIST~\cite{MNIST}, CIFAR-10~\cite{Cifar10}, ImageNet~\cite{ImageNet}, MS-COCO~\cite{MS-COCO} and many others. The data distribution on these sets was thoroughly checked based on many aspects, however the interesting, non-background pixels can usually be found in the middle of the images and thus objects are rarely present at the edges. On the other hand in case of certain applications --- such as collision prevention in self driving cars --- objects appearing at the edges of the images might be especially important.

Based on the architecture of convolutional networks one could assume that they are shift invariant, but unfortunately this common assumption is untrue and it was demonstrated recently \cite{zhang2019making} that these structures are dependent on spatial shifting. 

In this paper, based on the 2020 results of O. S. Keyhan and J. C. Gemert \cite{BoundaryEffectPaper} and the 2021 results of Md. Am. Islam \cite{PaddingArticle} we present evidence, that convolutional neural networks (CNNs) are capable of learning non translation equivariant behaviors --- which appears at the edges of the image ---, thus training and validating on the previously mentioned data sets might provide biased results and the trained CNNs might not reach the expected performance in realistic circumstances where objects may appear near the edges of the images. Furthermore we propose multiple validation environments capable of showing these biases in an emphasised manner and we propose multiple simple solutions to this issue which could prevent the biased behavior without any significant drawbacks besides a potentially increased training time.

\section{Object distribution on popular data sets} \label{sec:ObjectDistribution}

In the human vision system we have only one location, the fovea centralis in our eye, where the density of the light processing cones is significantly higher than the surrounding regions and this area is responsible for our detailed, central vision as well as main color vision. \cite{FoveaCentralis} Since we can focus typically on only one region it is easier for our nervous system to center the important elements in our field of view.
It is also more aesthetically pleasing to look at images where the important details and objects are at the center \cite{chen2018looking}.
This evolutionary consequence of human behaviour can introduce a significant bias in all human acquired datasets.

\begin{figure}
\centering
\subfloat{\includegraphics[width=.3\linewidth]{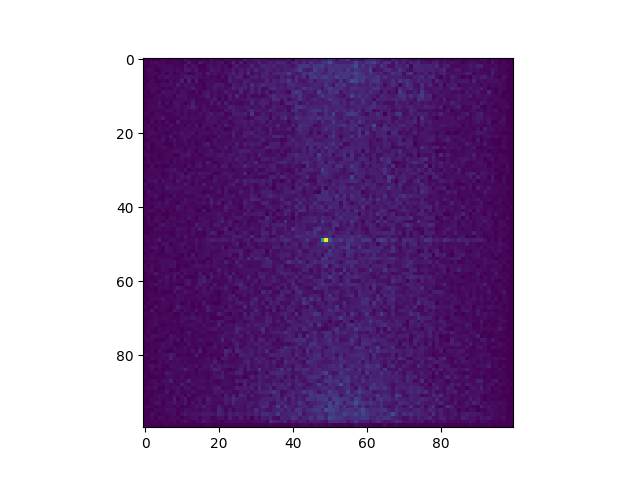}}
\hfill
\subfloat{\includegraphics[width=.3\linewidth]{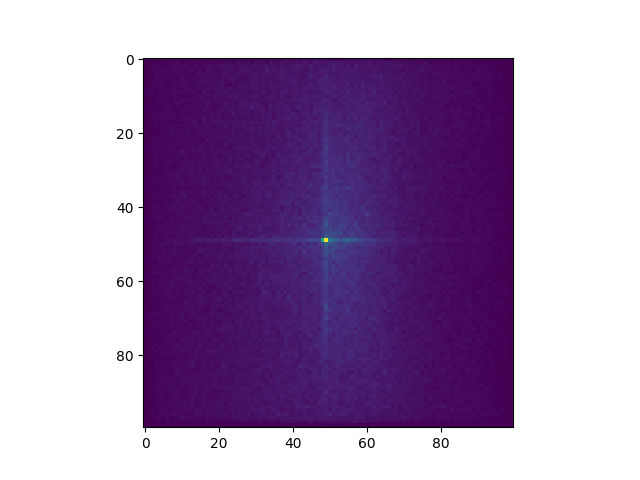}}
\hfill
\subfloat{\includegraphics[width=.297\linewidth]{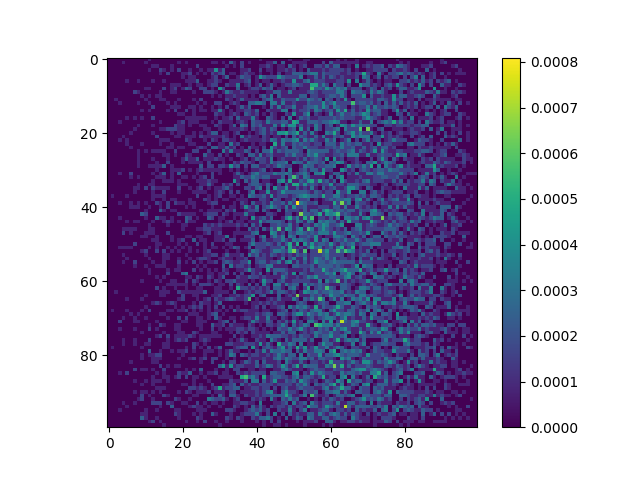}}
\hfill
\subfloat[Car]{\includegraphics[width=.28\linewidth]{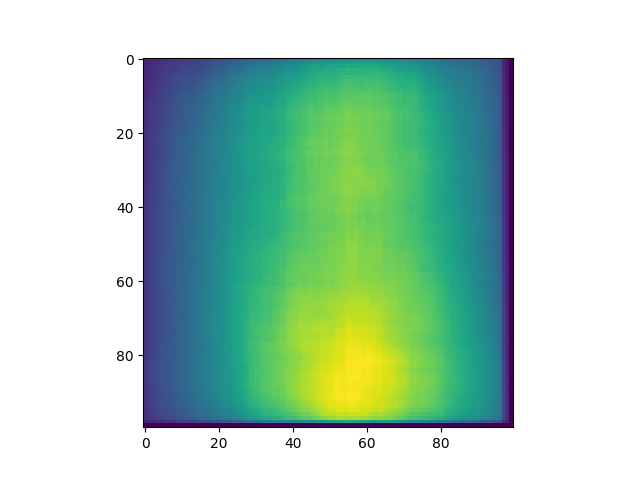}}
\hfill
\subfloat[Person]{\includegraphics[width=.28\linewidth]{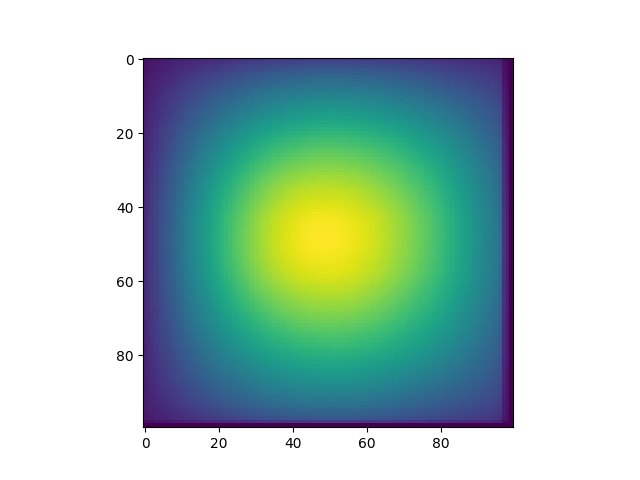}}
\hfill
\subfloat[Handbag]{\includegraphics[width=.28\linewidth]{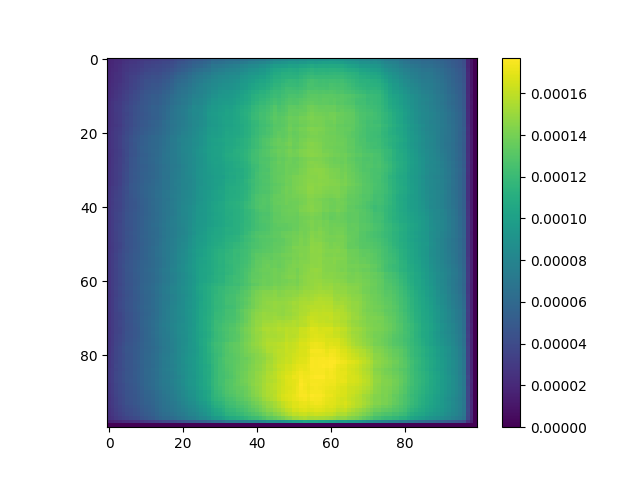}}
\caption{Appearance probability of certain objects on the MS-COCO dataset. In the top row the frequency of center points calculated for all objects are displayed as a heatmap, meanwhile in the bottom row the bounding boxes are displayed similarly. As one can see from the images, there is a high probability that the centroid or even the whole bounding box of a randomly selected object will be in the middle, meanwhile objects only seldomly appear around the edges of the image.}\label{FigCocoDist}
\end{figure}

This bias can be easily observed in simple datasets such as MNIST, where all the objects are centered and a completely black region can be found around the edges in all images. In other, more complex datasets such as CIFAR, ImageNet and MS-COCO this bias is still present, although it might not be so obvious.  It is difficult to measure this bias in case of classification datasets where the position of the objects are not annotated precisely, but can be easily computed in datasets where object locations are marked with bounding boxes or masks.
To illustrate this phenomenon some heatmaps representing the frequency of certain objects according to their positions in the MS-COCO dataset can be seen in Figure \ref{FigCocoDist}.

\captionsetup[subfigure]{labelformat=empty}

Of course one could assume that this bias is irrelevant in convolutional network training, since they are shift invariant, but as we have mentioned earlier and will demonstrate below, this phenomenon can introduce a significant bias in network accuracy, especially when objects are close to the boundary of the image.

\section{Regional training results on U-net architecture based CNNs}\label{sec:RegionalTraining}

\begin{figure}
\centering
\subfloat{\includegraphics[width=.3\linewidth]{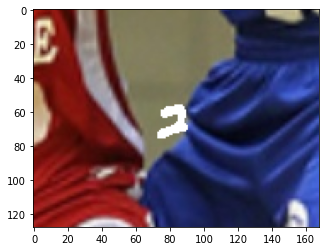}}
\hfill
\subfloat{\includegraphics[width=.3\linewidth]{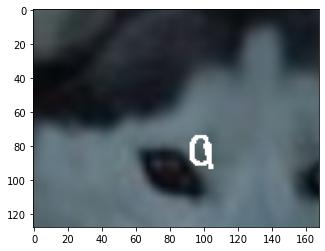}}
\hfill
\subfloat{\includegraphics[width=.3\linewidth]{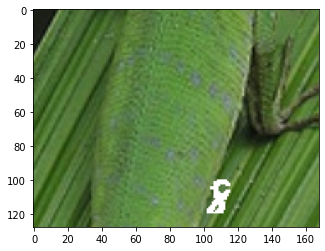}}
\hfill
\subfloat[Central]{\includegraphics[width=.3\linewidth]{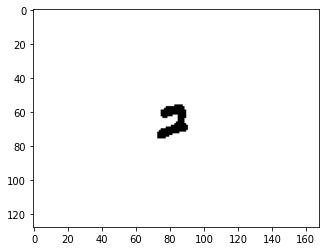}}
\hfill
\subfloat[Transitional]{\includegraphics[width=.3\linewidth]{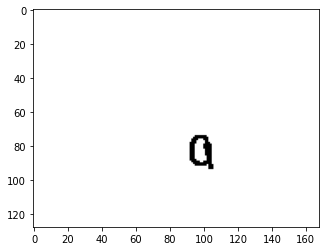}}
\hfill
\subfloat[Edge]{\includegraphics[width=.3\linewidth]{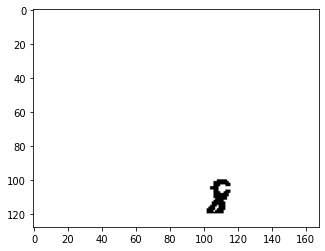}}
\caption{Example input images (upper row) and the corresponding background labels (lower row) for the trainings using ImageNet samples as backgrounds and MNIST handwritten digit samples as the objects to be segmented and classified.}\label{fig:InputImages}
\end{figure}

After realizing the possible non equivariant traslational behavior of CNNs during initial measurements, we designed a specific object segmentation task and evaluated it using the popular U-net architecture \cite{Unet}. In this section and in Section \ref{sec:SaliencyShiftMaps} all the measurement results were generated using the same U-net architecture, having cross entropy as the loss function, with 20 epochs of training on 60,000 randomly generated sample images with batch size of 32, and evaluated on 16,000 samples. The task itself was semantic segmentation and classification of handwritten digits using the MNIST handwritten digits data set, however we placed the images onto randomly selected and rescaled samples of the ImageNet dataset, serving as the background. Nonetheless the task was purposefully designed to be easy and definitely solvable for the U-net architecture, as the non-background regions always had the maximum possible value on the input images, and the classification of the MNIST handwritten digits data set is a trivial task even for non-state-of-the-art architectures. The handwritten digits were always positioned at the middle of the much larger ImageNet samples and we cropped a 168x128 region as the training and test samples. The limits of positioning were designed for all of the measurements mentioned in Section \ref{sec:RegionalTraining} and Section \ref{sec:SaliencyShiftMaps} in a way, that no parts of the objects left the image and in the most extreme cases the objects just touched an edge of the image. Such example input images with the corresponding labels can be seen in Figure \ref{fig:InputImages}.

\captionsetup[subfigure]{labelformat=empty}

Using these images, we performed 5-5 initial trainings with the same U-net architecture in order to test the hypotheses of non equivariant translation, where in half of the cases the handwritten digits were positioned at the central 30\% of the images (which is the central 9\% of the image in terms of area), and in the other half of the cases the handwritten digits were positioned outside the central 70\% of the images.

For the test data initially we created two sets in a similar manner and we measured the mean test losses for each neural network with both the similarly positioned test data, and the opposite. The loss values were almost the same on average for the similar data sets used at training and testing, however switching the data sets showed a remarkably extreme difference. Training only at the edges of the images and testing at the center increased the test loss values about 1.9 times on average, but training only at the center and testing at the edges increased the loss values more than 16000 times. The averaged values on the four possible pairings as well as the exact ratios between the losses for the same neural networks can be seen at Table \ref{tab:M2NIST_ImgNet_03-07Values}.   

\begin{table}
\resizebox{\linewidth}{!}{%
\begin{tabular}{llll}
\hline
 &
  \begin{tabular}[c]{@{}l@{}}Loss \\ central 30\% \end{tabular} &
  \begin{tabular}[c]{@{}l@{}}Loss \\excluded \\central 70\% \end{tabular} &
  \begin{tabular}[c]{@{}l@{}}Different/Same\\ loss ratio\end{tabular} \\ \midrule
\begin{tabular}[c]{@{}l@{}}Central training\end{tabular} &
  0.00041 &
  6.974354 &
  16,086.638 \\ 
\begin{tabular}[c]{@{}l@{}}Edge training\end{tabular} &
  0.000690 &
  0.000417 &
  1.898 \\ \hline
\end{tabular}
}
\caption{Mean values of the losses corresponding to the to the trainings and the evaluations at the center and at the edges, as well as the ratios between testing on the same type of data set versus testing on the opposite. As the ratios show, the results get worse in both cases if the training and test sets have different object placement patterns, however the increase of loss is far more drastic in case of central training.}\label{tab:M2NIST_ImgNet_03-07Values}
\end{table}

\begin{figure}
\centering
\subfloat{\includegraphics[width=.5\linewidth]{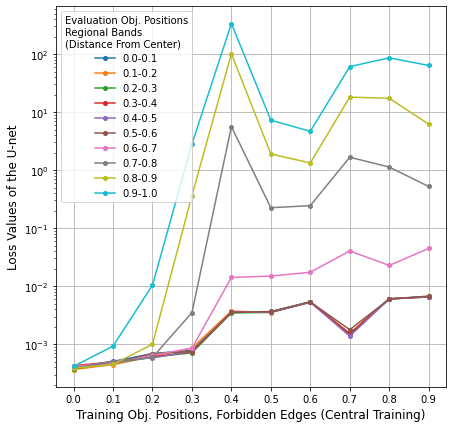}}
\hfill
\subfloat{\includegraphics[width=.5\linewidth]{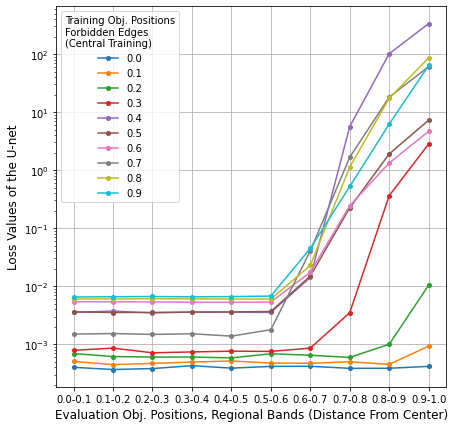}}
\hfill
\subfloat{\includegraphics[width=.5\linewidth]{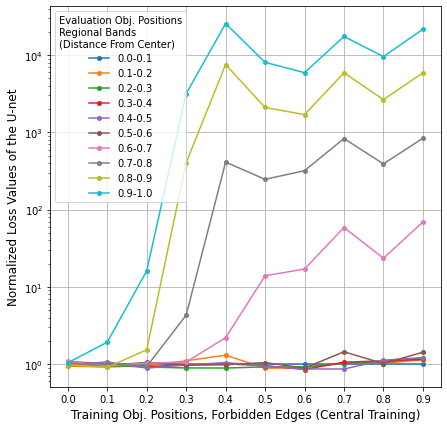}}
\hfill
\subfloat{\includegraphics[width=.5\linewidth]{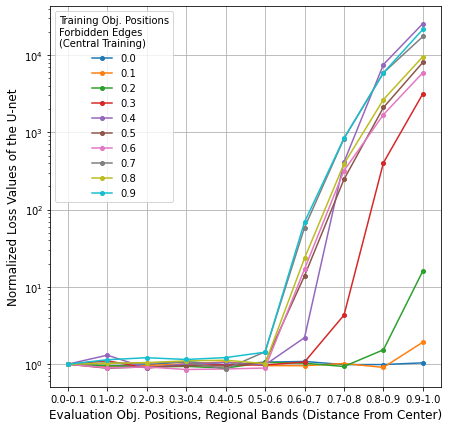}}
\hfill
\caption{Mean loss values of the U-nets trained on the "band regions" described in Section \ref{sec:RegionalTraining}. As it can be seen the loss values start to increase drastically beginning at the 0.5-0.6 evaluation band unless objects are specifically positioned near the edges during training, which means that in terms of area more than 64\% of an image will be segmented and classified with a significantly worse performance.}\label{fig:LossValuePlots_RegularConv}
\end{figure}

Following this initial measurement we localized the problematic region by decreasing the allowed central region on the training images for the objects from 1 to 0.1, and by placing the objects during the testing on bands from 0.0-0.1 to 0.9-0.1, where 0.0-0.1 marks the most central region of the image, while 0.9-1.0 marks the edge of the image. For each of the training cases we trained 5 independent neural networks with the same U-net architecture and averaged the results for each of the instances. The results --- which can be seen in Figure \ref{fig:LossValuePlots_RegularConv} --- show that when the objects are only placed in the central $60\%$ of the images, the mean losses start to increase rapidly even quite far (more than $30\%$ off) from the edges of the images. We emphasize that the objects are entirely present on the images even when they are placed at the $90\%$ band from the central region, so this drop can not be caused by occlusion. It can be seen as well, that the outer most region (0.9-1.0) is even sensitive to the objects being placed in the central $80\%$ of the images. Both the bit more restrictive central $60\%$ and the less restrictive central $80\%$ object placements seem to be very much realistic based on the measurement results described in Section \ref{sec:ObjectDistribution}, thus we conclude that unless special care is taken for the placement of the objects, the training and even the evaluation results on many popular data sets are destined to be biased, and the performance of the CNNs in realistic applications where objects may appear in a non-centered manner will decay significantly.

In all measurements we will list only the main and most determining parameters of our experiments. For the detailed set of parameters and implementation of all of our experiments using Pytorch \cite{Pytorch} please check our codes in the supplementary material.

\captionsetup[subfigure]{labelformat=empty}

\section{Saliency-shift maps of regionally trained U-nets} \label{sec:SaliencyShiftMaps}

\begin{figure}
\centering
\subfloat{\includegraphics[width=.3\linewidth]{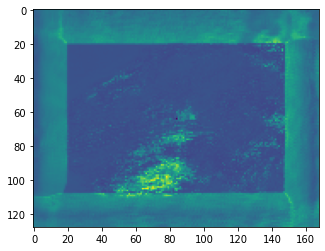}}
\hfill
\subfloat{\includegraphics[width=.3\linewidth]{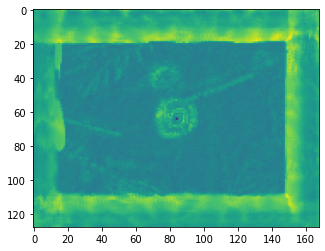}}
\hfill
\subfloat{\includegraphics[width=.35\linewidth]{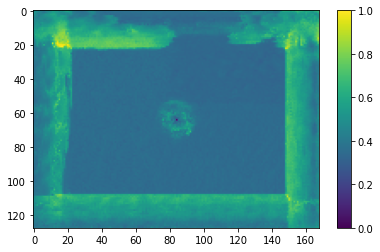}}
\hfill
\subfloat{\includegraphics[width=.3\linewidth]{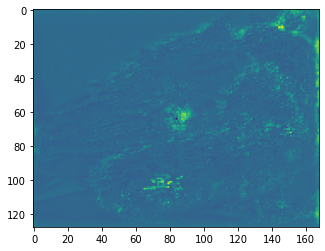}}
\hfill
\subfloat{\includegraphics[width=.3\linewidth]{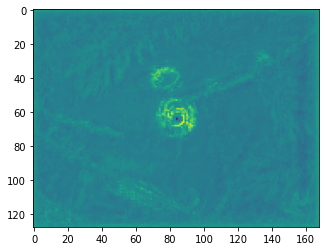}}
\hfill
\subfloat{\includegraphics[width=.35\linewidth]{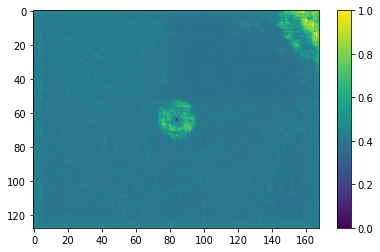}}
\caption{Normalized saliency map differences based on the shift vectors (zero shift positioned at the center). The front row depicts the results using U-nets trained with objects in the middle (central 30\% allowed), while the bottom row depicts the results using U-nets trained with objects at the edges  (central 70\% forbidden). As it can be seen, the CNNs trained on images with objects only at the center have a very distinct and large difference in their saliency maps when the objects are present at the edges, which the CNNs trained with objects at the edges do not have.}\label{fig:SaliencyDiffMaps_ImageNetBcgrnd}
\end{figure}

Following the initial measurements we hypothesized that depending on the position of the object the class-saliency maps \cite{SaliencyMaps} could change even if the position shift on the object was minuscule. In order to ascertain whether this information would lead to closer understanding on the phenomenons, we measured the difference of the saliency maps for setups  described in Section \ref{sec:RegionalTraining}. We measured the absolute differences between the saliency map generated on the non-shifted (centrally placed) object and the saliency maps generated on the same images, with the shifted object and background. Here it is important to mention again, that the objects have never left the image, in the most shifted cases the objects were still fully inside the image positioned near the edges. For better visualization we represented the saliency map differences as a matrix of same size as the input images with the coordinates representing the amount of shift in the given direction (the origin positioned at the center) and the values representing the dispersion-normalized \cite{DispersionNormalization} difference values. Some samples of resulting images can be seen in Figure \ref{fig:SaliencyDiffMaps_ImageNetBcgrnd}.

\captionsetup[subfigure]{labelformat=empty}

Based on these results we strongly suspect, that due to zero-padding having a categorically distinct effect --- which can not be achieved even by images having large completely black regions as the padding has zero values in each layer while the zero values originating from the input image can be subject to change from one layer to another --- we reckon that the effects of zero-padding (or possibly any other padding having characteristically different behavior as the input image itself) and the potentially large scale non equivariant translational behavior of the maximum pooling layers cause the outstanding differences of the loss values in different training circumstances described in Section \ref{sec:RegionalTraining}. Lastly we also suspect based on all the measurement results described in Section \ref{sec:RegionalTraining}, that there is a really high chance of a CNN learning unwanted, meaningless and destructive non equivariant translational behavior near the edges, but this can be easily prevented via deliberately positioning objects near the edges of the images during the training, as this object positioning forces the CNN not to take into count the otherwise unusual behavior of zero-padding.

We find some resemblance between the effect of boundary conditions and the mechanisms behind Dropout \cite{srivastava2014dropout} and especially Dropblock \cite{ghiasi2018dropblock}. We hypothesize that the zero-padding can be considered as a pathological case of Dropblock when systematically all activations are zerod out in every layer in a region. However, in case of Dropout and Dropblock the remaining activations are scaled to maintain the average activation while this mechanism is missing in case of boundary conditions, and the regions for Dropout and Dropblock are randomly chosen for each layer, while the boundary condition always affects the outermost pixels.

\section{Performance of Commonly Applied Networks at the Boundary}\label{sec:CommonNets}

\begin{figure}[htp]
 \centering
\subfloat{   \includegraphics[width=.2\linewidth]{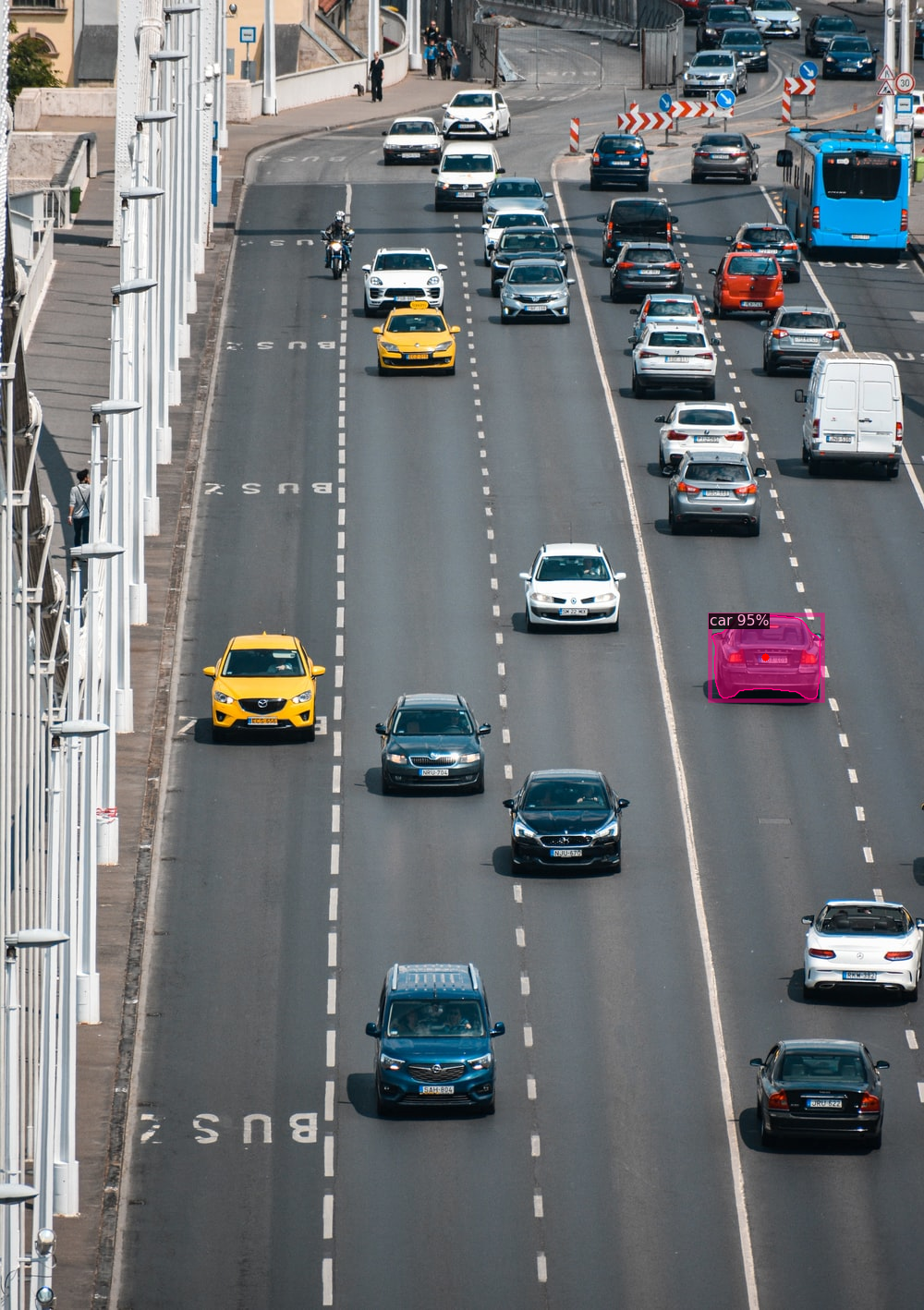}}
\subfloat{   \includegraphics[width=.6\linewidth]{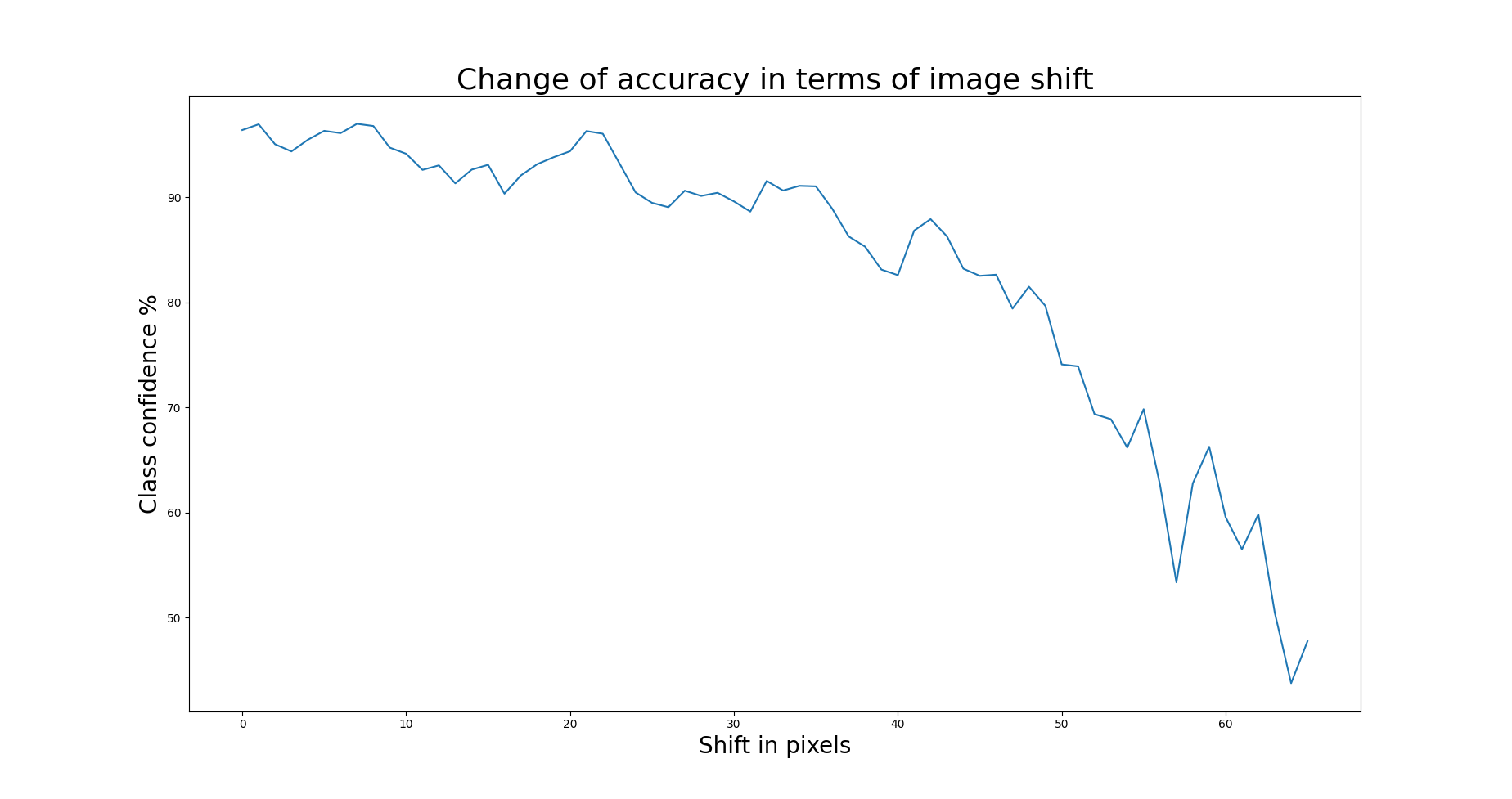}}
\caption{ Depiction of the boundary effect using a Mask R-CNN, with a ResNet-50 backbone and a feature pyramid network, pretrained on the MS-COCO dataset. The original input image can be seen on the left, where the object (marked by purple instance mask) is detected with high confidence. The right plot displays how confidence changes with each periodic right shifts of the image. Even in case of maximal shift none of the pixels of the object are out of frame, the drop in detection confidence happened only due to an indirect effect of the boundary. }
\label{fig:DetectronShift}
\end{figure}

To investigate the previously introduced caveat of neural networks we have investigated the Mask R-CNN architecture \cite{he2017mask}, implemented in the Detectron2 environment \cite{wu2019detectron2}. We have used a pretrained version of a Mask R-CNN network with ResNet-50 backbone \cite{he2016deep} with feature pyramid network and ROI align. For the sake of reproducibility this architecture was not trained sby us, but we have used the pretrained weights provided by Detectron 2 and investigated how object shift affect the detection accuracy of the network.
A similar drop in the objectness score of the detected objects was observed as they get closer and closer to the boundary of the image.
We note that this drop was unobserved for large objects which occupied more than $25\%$ of the image area. These objects were detected with high confidence even when they were partially out of frame. For example persons were detected even only when one of their arms was visible on the image. But the detection accuracy of smaller objects has dropped significantly as they approached the edge of the network. An image illustrating this phenomenon can be seen in Figure \ref{fig:DetectronShift}.

\begin{figure}[htp]
\subfloat[Agama 66\%]{   \includegraphics[width=.2\linewidth]{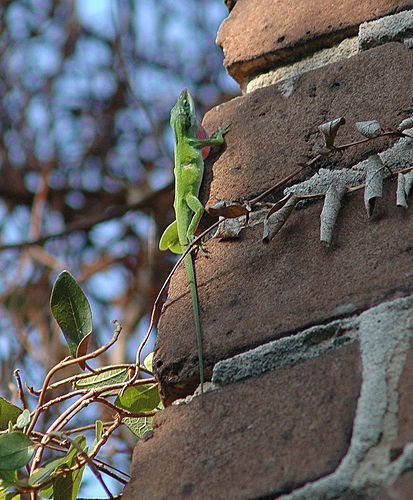}}
\hfill
\includegraphics[width=.4\linewidth]{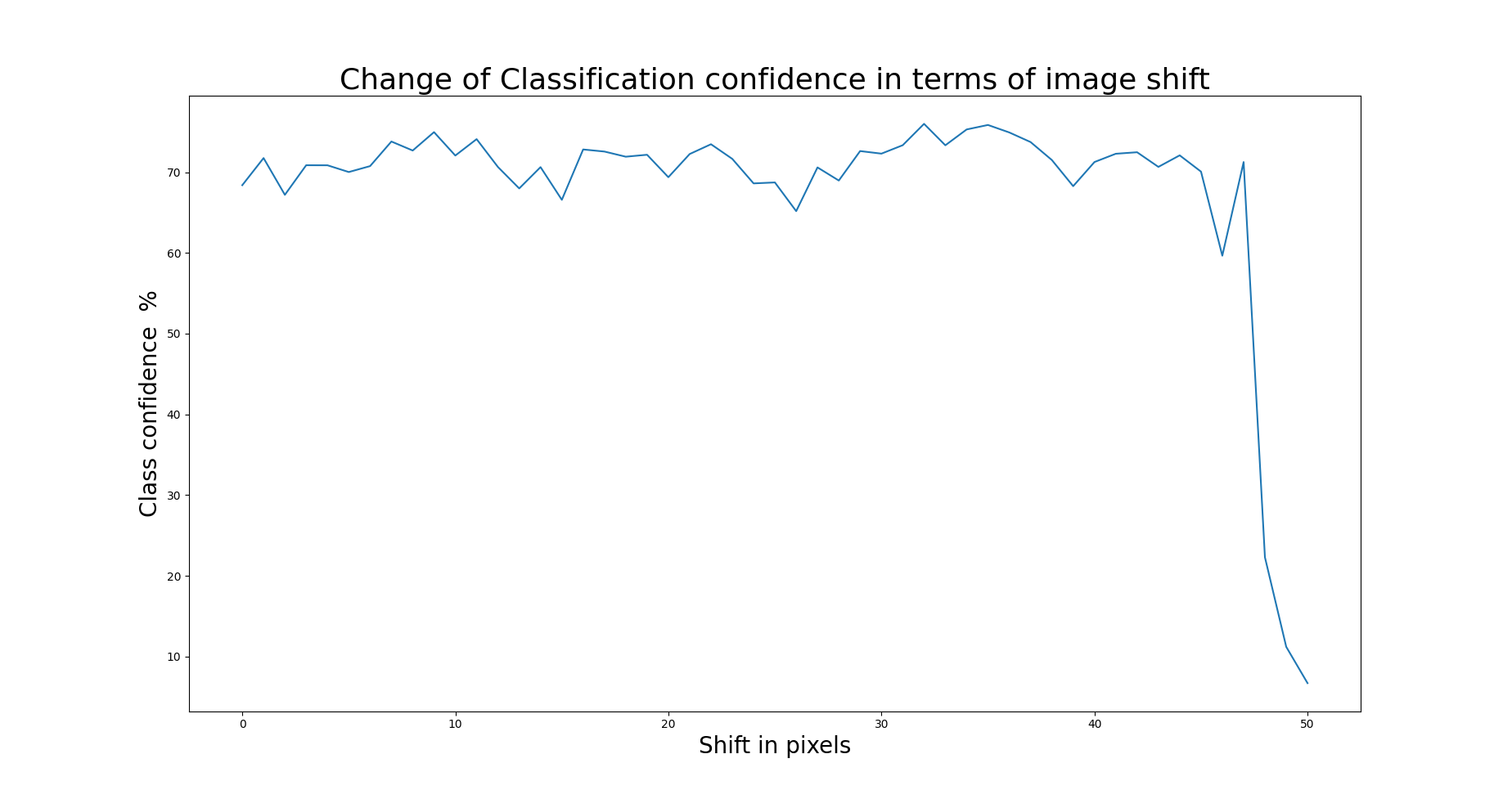}
\subfloat[Agama 6.7\%\\ Mud turtle 26\%]{   \includegraphics[width=.2\linewidth]{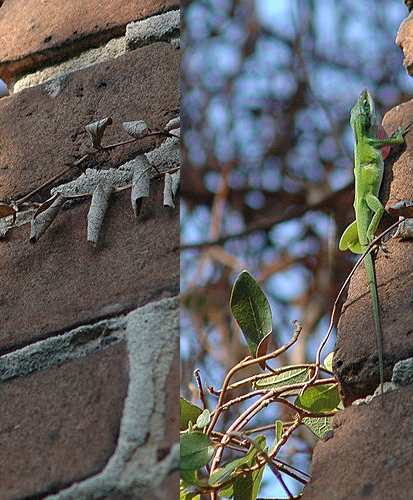}}
\\
\subfloat[Junco Snowbird 74\%]{  
\includegraphics[width=.2\linewidth]{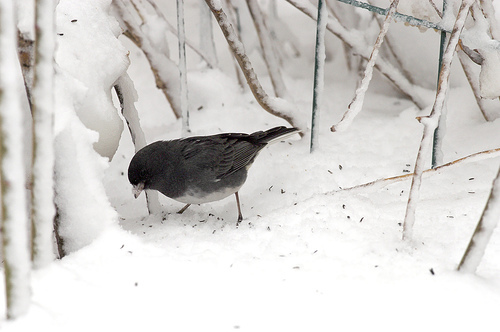}}
\hfill
\subfloat[]{
\includegraphics[width=.4\linewidth]{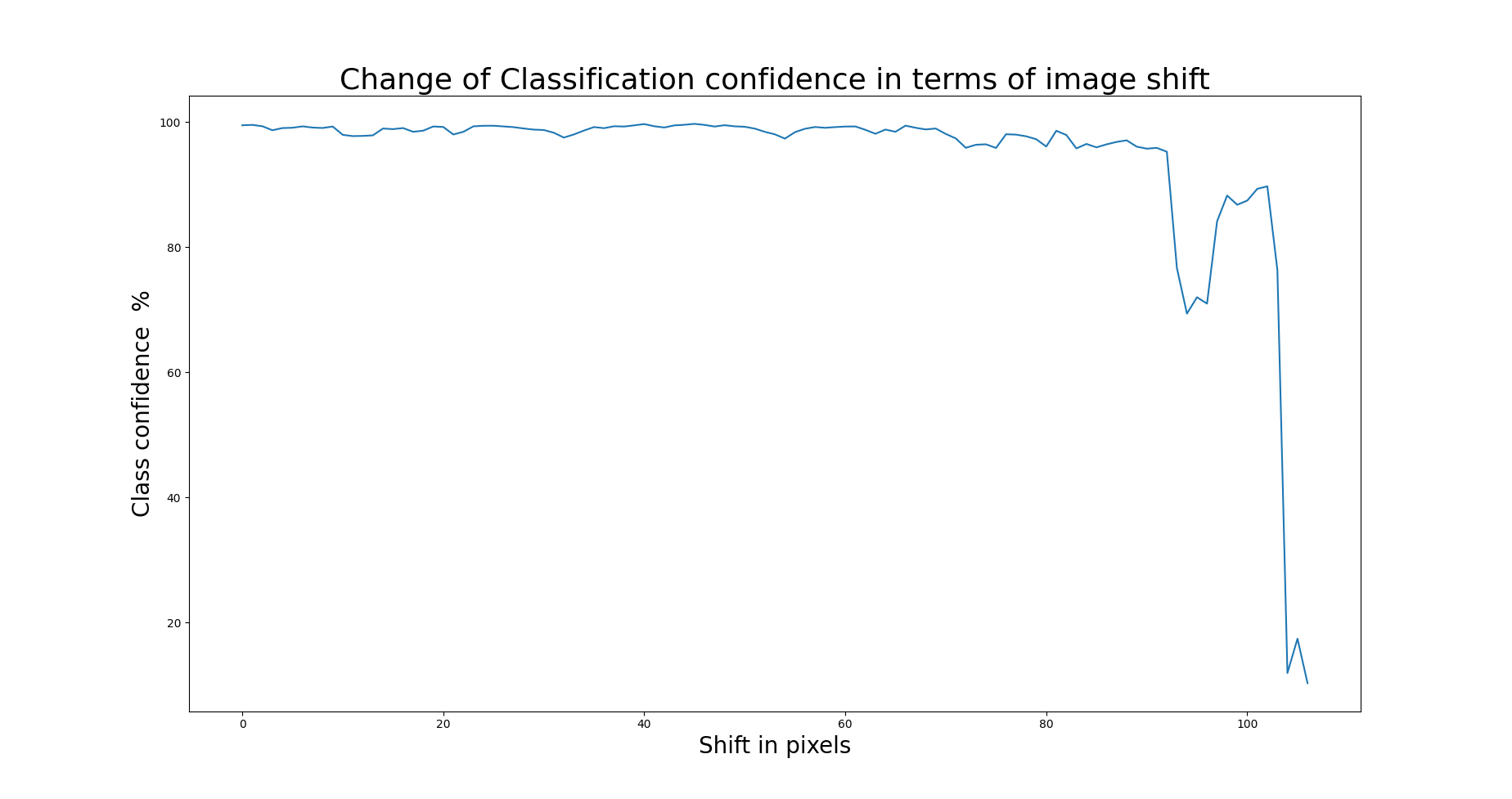}
}
\subfloat[Junco Snowbird 0.7\%\\ Ostrich 99\%]{   \includegraphics[width=.2\linewidth]{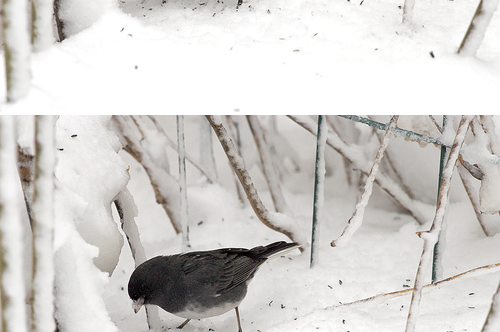}}
\caption{Sample cases showing the drastic effect of shifting objects to the boundary of the image on classification networks, using the pretrained version of VGG-16 with ImageNet samples. The first column display the original images which are classified correctly. The second column depicts how the classification confidence of the original class changes by shifting the image. The last column depicts the largest investigated shift. The classification confidence for the original class and the winning class are displayed below the images along with their confidence scores.}
\label{fig:ImageNetShift}
\end{figure}

We have also examined the performance of commonly applied classification networks, such as VGG-16 on the ImageNet dataset.
We have selected 10 classes and 10 images from each class. We have manually shifted these 100 images to create a dataset where the interesting objects can always be found at the boundary.
We have downloaded pretrained models from the model library of TorchVision \cite{marcel2010torchvision} and investigated their performance.
These models have high classification accuracy and were working well on the original samples but the performance has dropped significantly and usually caused misclassification when objects got close to the boundary.
An image depicting this effect can be seen in Figure \ref{fig:ImageNetShift}.
These plots depict that the shift invariance causes only a minor change in classification accuracy, but when the object gets close to the boundary an abrupt decrease can be observed, which is not related to shift invariance.
We emphasize that the periodic shift we have applied can introduce a strong edge at the middle of the image, but this was not the reason for the accuracy drop, since the classification result changed only slightly in case of smaller shifts, which also introduced this effect.
As it can be seen from these results the phenomenon described and investigated in the previous section via simple datasets also exists in case of complex networks and problems, including not only classifcation but detection and segmentation as well.

\section{Solutions for the Mitigation of the Object Centering Bias}\label{sec:Solutions}

As the results in Section \ref{sec:RegionalTraining} demonstrates, having objects solely in the central regions during training can introduce a significant bias to CNNs. On the other hand while training with objects solely at the edges of the images still introduces a certain level of bias, its effect will be lower by multiple orders of magnitude as in the opposite case.
In this section our aim is to explore some data-manipulation based and solely architectural solutions to mitigate the significant effect of object centering bias.

\subsection{Image cropping}\label{sec:ImageCropping}

One simple and obvious solution of this problem could be the neglection of the outer regions of images. With this method one could execute a network with smaller input dimensions, without using any boundary conditions at all and using the unused pixels during the computation. This method can fix the object centering bias in a straight forward way, but unfortunately it would also decrease the visible and investigated region by the network, which may not be admissible in many applications. Also in order to perform such image cropping based transformations, the shape and the position of the objects must be known, otherwise the cropping could frequently result in the loss of features or even entire objects. As a result of these issues, this method is unusable for a significant amount of real-world applications, while the other solutions we describe here do not suffer from this issue.

\subsection{Image shifting}\label{sec:ImageShifting}

If the application and the available data allows based on the measurement results in Section \ref{sec:RegionalTraining} we believe that changing or augmenting the training and evaluation data sets to have a uniform spatial distribution for every object class could be one of the best solutions.

To validate this hypothesis we have retrained a Mask R-CNN on the MS-COCO dataset where objects were shifted to the boundary of the image. In case of multiple objects, which is typical on this dataset, one object was selected randomly and the image was shifted periodically towards the closest boundary of the object with an amount of shift which ensured that the edge of the bounding box is exactly at the boundary.

Similarly we have retrained different classification network architectures on ImageNet where we have applied a large random translation on the images. Translation is typically applied in data augmentation, but usually images are shifted only by a few pixels. We have used periodic translations with a random amount which were ranging between zero pixels and one quarter of the width and height of the image. This can introduce a lot of problems and artifacts in the images (for example an object can be cut in half), since the exact location of the objects is unknown in this dataset, on the other hand the object will be more likely to be close to the boundaries.

Once these networks were trained we have examined their performance on the original dataset and two modified datasets where the objects were moved to the boundary. We have used our earlier described  small ImageNet subsample, containing hundred images with objects at the boundary and in case of MS-COCO we used the previously described method to generate randomly shifted samples for the test images.
The results on the ImageNet dataset can be seen in Table \ref{tab:ImageNEt} meanwhile the results for object detection are displayed in Table \ref{tab:MSCOCO}.
As it can be seen from these results the network performance has dropped slightly on the original test sets, when objects appeared at the boundary of the image during training. We assume that this can be caused by the artifacts introduced by our shifts.
But on the test sets where objects appeared at the boundary, the networks where the central bias appeared during training performed horribly. Their top-1 accuracy has dropped by more than $65\%$ in case of classification and around $15\%$ in detection and segmentation problems.
Meanwhile the networks where objects were shifted to the boundary during training (or more towards the boundary randomly in case of classification) 
the accuracy dropped only slightly on the original dataset ($4\%$ in average) and their performance were comparable on the shifted test sets.

In comparison to the image cropping solution briefly described in Section \ref{sec:ImageCropping}, this image shifting method is usable for more applications. On the other hand we also believe that architectural solutions --- like the one we will describe in the following subsection --- could be objectively better and easier to use as it will not require the manipulation of the training data set, would avoid the issues with newly introduced artefacts, and they would work for all data sets even if the shape and position of the objects is unknown.

\begin{table}[hpt!]
\centering
\resizebox{\linewidth}{!}{%
\begin{tabular}{@{}lccc@{}}
\hline
   \textbf{Architecture} & ImageNet & ImageNet Boundary \\ \midrule
\hline

VGG-16 Orig&  $70.8\%$ &     $7\%$    \\
VGG-16 Shifted&  $66.4\%$ &     $62\%$   \\
VGG-16 Toroidal&  $70.6\%$ &     $66\%$   \\
ResNet-50 Orig&  $72.1\%$ &     $5\%$      \\
ResNet-50 Shifted&  $66.75\%$ &     $61\%$     \\
ResNet-50 Toroidal&  $71.8\%$ &     $65\%$     \\
DenseNet121 Orig&  $75.1\%$  &    $3\%$     \\
DenseNet121 Shifted&  $69.3\%$   &    $66\%$    \\
DenseNet121 Toroidal&  $74.2\%$ &     $68\%$     \\
\hline
\end{tabular}}
\caption{Top-1 test accuracies of different architectures on the ImageNet test set (first column) and our manually created small subsample, where images are at the boundary (second column). The rows contain three different architectures: VGG-16 \cite{simonyan2014very}, ResNet-50 \cite{he2016deep} and DenseNet-121 \cite{huang2017densely}. Every network has three variants: trained on the original ImageNet training set with zero-padding (Orig.), the same architecture trained on the shifted version of the dataset (Shifted) and trained on the original dataset, but with toroidal boundary conditions (Toroidal). }
\label{tab:ImageNEt}
\end{table}
\vspace{-10pt}

\begin{table}[hpt!]
\centering
\resizebox{\linewidth}{!}{%
\begin{tabular}{@{}lccc@{}}
\hline
   \textbf{} & MS-COCO Orig & MS-COCO Shifted\\ \midrule
\hline
Box mAP Orig &  $33.6\%$ &     $15.6\%$    \\
Box mAP Shifted &  $31.7\%$ &     $28.4\%$    \\
Seg mAP Orig&  $31.4\%$ &     $17.4\%$    \\
Seg mAP Shifted&  $29.4\%$ &     $25.3\%$    \\
\hline
\end{tabular}}
\caption{Summary of the mean average precision results on the Ms-COCO dataset for a Mask R-CNN network with a ResNet-50 backbone and a feature pyramid network with ROI align. The two columns contain results for two versions of the test set of MS-COCO dataset: the original (MS-COCO Orig) and a version where we always shifted a randomly selected object to the boundary (MS-COCO Shifted). We have measured the mean average precision of the bounding box detection (Box) and instance segmentation tasks (Seg) of two different trainings and these are displayed in the rows. The rows marked with Orig display the results on unaltered Coco dataset, meanwhile the rows marked with shifted contain mAPs when the network was trained on the dataset where a randomly selected object was always shifted to the boundary.}
\label{tab:MSCOCO}
\vspace{-5mm}
\end{table}

\subsection{Toroidal boundary condition}

Toroidal or other periodic boundary conditions are widely used in classical image processing, physics simulations, cellular automatons and other fields, however in convolutional neural networks their usage is not a commonly applied technique. Based on the results of the image shifting solution, which we described at Section \ref{sec:ImageShifting}, we made the educated guess that using toroidal boundary condition for the convolutional layers could solve the object centering bias in a similar manner --- using images with no artificially created edges as in case of zero-padding --- but as an architectural solution.

Furthermore in case of only using shifted images all the hidden layers of the network still have to deal with the effects of zero-padding. Meanwhile in case of toroidal boundary condition for each of the convolutional layers, the network never has to handle the outlying nature of fully zero boundary conditions and the influx of these zero values --- which would still appear for the hidden layers even in case of images with black edges.

\begin{figure}
\centering
\subfloat{\includegraphics[width=.3\linewidth]{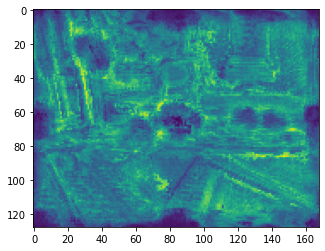}}
\hfill
\subfloat{\includegraphics[width=.3\linewidth]{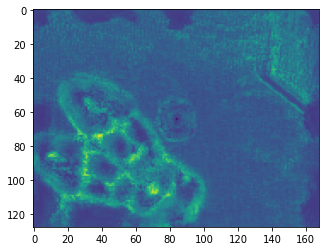}}
\hfill
\subfloat{\includegraphics[width=.35\linewidth]{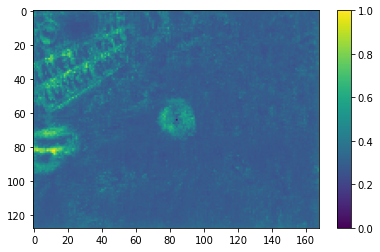}}
\hfill
\subfloat{\includegraphics[width=.3\linewidth]{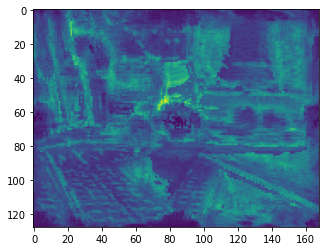}}
\hfill
\subfloat{\includegraphics[width=.3\linewidth]{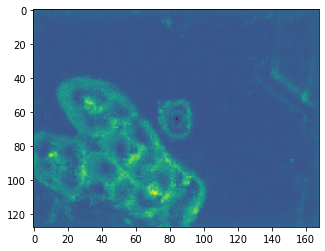}}
\hfill
\subfloat{\includegraphics[width=.35\linewidth]{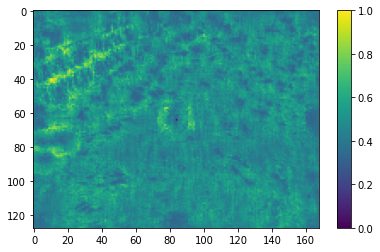}}

\caption{
Normalized saliency map differences similarly calculated and displayed as in case of Figure \ref{fig:SaliencyDiffMaps_ImageNetBcgrnd} (top row: central 30\% allowed, bottom row: central 70\% forbidden during training), but using toroidal boundary condition. As it can be seen, the large differences at the edges are completely gone.
}\label{fig:SaliencyDiffMaps_ImageNetBcgrnd_Periodic}
\vspace{-8mm}
\end{figure}

To measure the effect of toroidal boundary condition we have performed the same regional-training and evaluation measurements we described in Section \ref{sec:RegionalTraining}, but using toroidal boundary condition for the convolutional layers. The loss values for each training region versus each evaluation band can be seen in Table \ref{tab:M2NIST_ImgNet_ArMeanLosses_Bands_Toroidal}. As these results show, the usage of periodic boundary condition for each convolutional layer almost completely negates the effect of the object centering bias. Comparing the loss values between zero-padding and toroidal boundary, in the outermost cases toroidal boundary performs more than 37000 times better than zero-padding without any significant effect on accuracy in the central regions.
We have also measured the effect of toroidal boundary condition on ImageNet and the shifted version of this dataset where objects are at the boundary. Our results can be seen in Table \ref{tab:ImageNEt} and they clearly demonstrate that changing the boundary condition can improve the networks poor performance at the boundary.
Furthermore we have measured the difference of saliency maps based on shifting similarly as we described at Section \ref{sec:SaliencyShiftMaps}, and as it is clearly visible from the comparison of the results displayed in Figure \ref{fig:SaliencyDiffMaps_ImageNetBcgrnd} and Figure \ref{fig:SaliencyDiffMaps_ImageNetBcgrnd_Periodic}, training with only objects at the center can provide fundamentally the same results as training at the edges, which was not the case with zero-padding. Furthermore all the larger differences also got eliminated this way, and the images at Figure \ref{fig:SaliencyDiffMaps_ImageNetBcgrnd_Periodic} only show the smaller perturbations caused by the background of the individual test images scaled up by normalization.

\begin{table}[hpt!]
\resizebox{\linewidth}{!}{%
\begin{tabular}{lrrrrr}
\hline
\begin{tabular}[c]{@{}l@{}}Forbidden \\ Outer \\ Region\end{tabular} &
  \multicolumn{1}{l}{\begin{tabular}[c]{@{}l@{}}LossVal\\ 0.0-0.1\end{tabular}} &
  \multicolumn{1}{l}{\begin{tabular}[c]{@{}l@{}}LossVal \\ 0.1-0.2\end{tabular}} &
  \multicolumn{1}{l}{\begin{tabular}[c]{@{}l@{}}LossVal \\ 0.2-0.3\end{tabular}} &
  \multicolumn{1}{l}{\begin{tabular}[c]{@{}l@{}}LossVal \\ 0.3-0.4\end{tabular}} &
  \multicolumn{1}{l}{\begin{tabular}[c]{@{}l@{}}LossVal \\ 0.4-0.5\end{tabular}} \\ \hline
0.0 & 0.00038 & 0.00038 & 0.00039 & 0.00033 & 0.00042 \\
0.1 & 0.00049 & 0.00055 & 0.00053 & 0.00046 & 0.00052 \\
0.2 & 0.00053 & 0.00060 & 0.00065 & 0.00058 & 0.00063 \\
0.3 & 0.00075 & 0.00073 & 0.00067 & 0.00079 & 0.00078 \\
0.4 & 0.00069 & 0.00072 & 0.00060 & 0.00053 & 0.00082 \\
0.5 & 0.00069 & 0.00062 & 0.00082 & 0.00073 & 0.00075 \\
0.6 & 0.00074 & 0.00068 & 0.00052 & 0.00061 & 0.00057 \\
0.7 & 0.00062 & 0.00063 & 0.00065 & 0.00066 & 0.00053 \\
0.8 & 0.00082 & 0.00069 & 0.00072 & 0.00074 & 0.00062 \\
0.9 & 0.00064 & 0.00067 & 0.00059 & 0.00057 & 0.00081 \\ \midrule\hline
  \multicolumn{1}{l}{\begin{tabular}[c]{@{}l@{}}\vspace{18pt}\end{tabular}}&
  \multicolumn{1}{l}{\begin{tabular}[c]{@{}l@{}}LossVal \\ 0.5-0.6\end{tabular}} &
  \multicolumn{1}{l}{\begin{tabular}[c]{@{}l@{}}LossVal \\ 0.6-0.7\end{tabular}} &
  \multicolumn{1}{l}{\begin{tabular}[c]{@{}l@{}}LossVal \\ 0.7-0.8\end{tabular}} &
  \multicolumn{1}{l}{\begin{tabular}[c]{@{}l@{}}LossVal \\ 0.8-0.9\end{tabular}} &
  \multicolumn{1}{l}{\begin{tabular}[c]{@{}l@{}}LossVal \\ 0.9-1.0\end{tabular}} \\ \hline
0.0 & 0.00042 & 0.00041 & 0.00037 & 0.00048 & 0.00041 \\
0.1 & 0.00050 & 0.00049 & 0.00052 & 0.00053 & 0.00057 \\
0.2 & 0.00060 & 0.00050 & 0.00074 & 0.00062 & 0.00063 \\
0.3 & 0.00087 & 0.00064 & 0.00066 & 0.00091 & 0.00114 \\
0.4 & 0.00073 & 0.00069 & 0.00075 & 0.00072 & 0.00150 \\
0.5 & 0.00078 & 0.00083 & 0.00086 & 0.00074 & 0.00128 \\
0.6 & 0.00076 & 0.00069 & 0.00076 & 0.00077 & 0.00090 \\
0.7 & 0.00059 & 0.00051 & 0.00066 & 0.00069 & 0.00156 \\
0.8 & 0.00059 & 0.00085 & 0.00077 & 0.00098 & 0.00164 \\
0.9 & 0.00064 & 0.00074 & 0.00081 & 0.00128 & 0.00171 \\ \hline
\end{tabular}
}
\caption{Means of losses of training-evaluation pairs in each table row on 5 individually trained U-nets utilizing toroidal boundary condition for the convolutional layers. The rows represent the loss values corresponding to the same set of CNNs trained on images with a given "outer" restriction on the object position. The columns represent the normalized loss values corresponding to the same regional band restrictions on the object position during evaluation.}\label{tab:M2NIST_ImgNet_ArMeanLosses_Bands_Toroidal}
\vspace{-8mm}
\end{table}

\captionsetup[subfigure]{labelformat=empty}

\section{Conclusion}

In this paper we have demonstrated an intriguing failing of convolutional networks. Because boundary conditions introduce zero activations on each layer, objects appearing close to the edge of an image will yield drastically different activations than at the center of the image.
Since most datasets have a bias centering the objects, convolutional networks trained on commonly used datasets perform poorly if objects are present at the boundary.
We have thoroughly investigated this phenomenon on classification and instance segmentation tasks on the MNIST, ImageNet and MS-COCO datasets.
We have also shown how this problem can be alleviated by shifting the image to the boundary during training or applying toroidal boundary conditions instead of zero-padding.
This training methodology has significantly increased the accuracy of the network at the boundary, which might be important in several practical applications.

\bibliographystyle{ieeetr}
\bibliography{edge.bib}

\end{document}